\documentclass[sigconf]{acmart}
\usepackage{enumitem}
\usepackage{subcaption}

\AtBeginDocument{%
  }

\begin{document}

%\title[A Synthetic Human Touch]{A Synthetic Human Touch: Modeling and Delivering Group Recommendations via Subjective Judgment and LLMs} % Fine-Tuned LLMs?
\title[Consensus vs. Dissent]{ Consensus vs. Dissent: Dynamic LLM Modeling of Subjective Preferences in Group Recommenders}
\author{Cedric Waterschoot}
\affiliation{%
  \institution{Maastricht University}
  \city{Maastricht}
  \country{The Netherlands}}
\email{cedric.waterschoot}
\email{@maastrichtuniversity.nl}

\author{Nava Tintarev}
\affiliation{%
  \institution{Maastricht University}
  \city{Maastricht}
  \country{The Netherlands}}
\email{n.tintarev}
\email{@maastrichtuniversity.nl}

\author{Francesco Barile}
\affiliation{%
  \institution{Maastricht University}
  \city{Maastricht}
  \country{The Netherlands}}
\email{f.barile}
\email{@maastrichtuniversity.nl}

\renewcommand{\shortauthors}{Waterschoot et al.}

\begin{abstract}
Previous work in group recommender systems has demonstrated a sensitivity to the distribution of preferences within a group. Specifically, the selection of the preference aggregation strategy benefits from considering such group configurations. 
In this paper, we study whether LLMs are able to mimic this sensitivity and to select the ideal aggregation strategy (and corresponding recommendation) according to nuanced human perceptions of fairness, satisfaction, and consensus.

We do this by fine-tuning Large Language Models (LLMs) on human survey data to serve as real-time judgmental models within the recommendation pipeline. Using a reasoning dataset distilled from DeepSeek-V3.1 and human ground truth assessments, we develop Judgmental Llama and Judgmental OLMo to simulate group assessments. Our pipeline successfully generates multiple recommendation candidates based on social choice-based aggregation strategies and dynamically selects the one that maximizes these predicted human-like evaluations. We further validate these suggestions in a user study ($n=284$) and find that our methodology achieved the highest scores for satisfaction and group consensus. Furthermore, we find that LLM judgments are most aligned with human perceptions of fairness, satisfaction and consensus when we also consider interaction effects between our LLM-based method and group configuration (e.g., minority or coalition). These findings give further support for dynamically adapting aggregation strategies to specific within-group preference distributions, and highlight the advantage of using LLMs for an adaptation that is aligned with subjective human judgments. %to move beyond static formulas toward GRS that align with subjective human assessments.

\end{abstract}


%% CCS TAGS



\maketitle

\section{Introduction}
\begin{comment}
problem statement:
Difficulty adapting group recommendations to group itself. User studies to exist tha ask participants to provide subjective assessments of perceived satisfaction, fairness and group consensus. However, these are simply user studies and cannot be implemented in an actual recommendation pipeline. In our paper, we use such user study (survey) data to finetune LLMs which can deliver the human assessments. We implement these in the recommendaiton pipeline, meaning that they provide human-like assessements of fairness, satisfaction and consensus of several recommendation options provided by different social-choice based aggregation strategies. Our recommendation pipeline then provides the best scoring group recommendation to the group. We evaluate our approach with a user study in which participants either rate a static social choice-based aggregation strategy or our dynamic approach based on fine-tuned LLMs.
\end{comment}

The challenge for Group Recommender Systems (GRS) is the inherent difficulty of generating and adapting recommendations to the nuanced, sometimes conflicting, preferences of distinct group members. GRS research has often relied on social choice-based aggregation strategies, such as Average, Least Misery, or Most Pleasure, to bridge the gap between individual ratings and group outcomes \cite{senot2010analysis}. While these methods are computationally efficient, transparent, and well-evaluated in the literature, they are essentially static mathematical calculations. To evaluate their effectiveness, researchers typically conduct user studies to gather subjective assessments of perceived satisfaction, fairness, and group consensus from human participants \cite{barile2023evaluating,BARILE2026Critical}.

However, a significant gap exists between evaluation and implementation. While user studies provide invaluable insights into how recommendations may be perceived by a group, these subjective metrics cannot be easily integrated into an automated, real-time recommendation pipeline. Current systems lack a scalable mechanism to predict how a specific group will feel about a recommendation before it is presented. As a result, GRS are not adaptable to account for the nuanced group scenarios and dynamics that human participants account for during a user evaluation.

In this paper, we propose a novel approach that bridges this gap by leveraging Large Language Models (LLMs) and the survey data in the literature. We utilize existing human data to fine-tune LLMs, transforming them into \textit{judgmental models} capable of delivering human-like evaluations. By integrating these models directly into a recommendation pipeline, we move beyond static aggregation. More precisely, our system generates multiple recommendation options via various social choice-based strategies and employs our fine-tuned models to generate multiple assessments per outcome, mimicking a group of human participants evaluating the specific group recommendation. Afterwards, our system selects the outcome that maximizes these human-like assessments, providing the group with a final recommendation which was perceived as fair, satisfactory and as group consensus for their specific preference profile. We evaluated our approach through a user study, measuring our dynamic, LLM-based method against traditional static strategies. Our user evaluation showed that our LLM-based selection of aggregation strategy achieved the highest average scores for perceived fairness and consensus, as well as the second highest for perceived satisfaction. All in all, our results show that our dynamic system is capable of managing both polar opposites: groups with high consensus (uniform) and groups with clear dissent (coalitional). Full documentation, code and data are available in the companion repository: \url{https://github.com/Cwaterschoot/Consensus-vs-dissent-2026/tree/main}

\noindent In short, our paper makes the following contributions:

\begin{itemize}[leftmargin=*]
    \item We present a distilled \textbf{reasoning dataset} based on DeepSeek-V3.1, grounded in human assessments of fairness, satisfaction, and consensus from \citet{barile2023evaluating,barile2023data}.
    \item We introduce \textbf{Judgmental Llama and Judgmental OLMo}, models specifically fine-tuned and evaluated on human judgment and distilled reasoning to simulate group assessments.
    \item We develop an end-to-end \textbf{recommendation pipeline} that employs these fine-tuned models to dynamically select group recommendation outcomes based on human-like evaluations.
    \item We conduct a user study ($n=284$) to evaluate the effectiveness of our judgment-based GRS and show that our proposed method achieved the highest assessments for fairness and consensus. Additionally, our user study showcases statistically significant interactions with group configuration.
\end{itemize}

\begin{table*}
  \caption{Preference aggregation strategies used in this study \cite{tran2019towards,barile2023evaluating,felfernig2018explanations}}
  \label{tab:strats}
  \begin{tabular}{ll}
    \toprule
    Strategy & Procedure \\
    \midrule
    Additive Utilitarian (ADD) & Recommends the item with the highest sum of all group members’ ratings \\
    Fairness (FAI)&  Ranking and recommending items according to how individuals choose them in turn\\
    %\midrule
    Approval Voting (APP)  &  Recommends the item with the highest number of ratings above a predefined threshold\\
   % \midrule
    Least Misery (LMS)  & Recommends the item which has the highest of all lowest per-item
ratings\\
    Most Pleasure (MPL)  & Recommends the item with the
highest individual group member rating\\
    Majority (MAJ)  &Recommends the item with the highest
    number of all ratings representing the majority \\
    &  of item-specific ratings\\
    \bottomrule
  \end{tabular}
\end{table*}

\section{Related Work}\label{sec:background}
In the following paragraphs, we outline the related work on GRS and the use of LLMs within the context of group recommendation. Additionally, we discuss fairness in (group) recommendation and how it is evaluated in the literature. Finally, we present the literature on using LLMs to automate the evaluation of such metrics under the paradigm of LLM-as-judge.

\subsection{Group Recommendation}\label{sec:rel-GRS}
% the usual about SCBAS, group configurations, the uses of GRS
Traditional recommender systems generate a recommendation on the individual level. In contrast, GRS need to process multiple, possible conflicting, preferences to produce a group recommendation which reflects the preferences of the group members \cite{Masthoff2022group}. We find recent applications for GRS in restaurant recommendation (e.g., \cite{barile2021toward,barile2023evaluating,Waterschoot2025withfriends}, tourism~\cite{chen2021attentive,delic2024supporting}, content moderation \cite{Waterschoot2024time}, or music recommendation~\cite{najafian2018generating}.

A large variety of methodologies have been proposed to generate such group recommendations based on a set of individual preference ratings. \citet{cao2018attentive} and \citet{huang2020efficient} put forward GRS based on attentive neural networks, while others have proposed graph neural networks \cite{Zhang2021double} or reinforcement learning \cite{Stratigi2023squirrel}. While these recent methodologies present promising avenues for generating group recommendations, social choice-based aggregation strategies, rooted in \textit{Social Choice Theory} \cite{kelly2013social}, remain a prominent procedure in the GRS literature, whether as main aggregation step~\cite{barile2023evaluating,tran2019towards,Waterschoot2025Pitfalls} or baseline in comparison to more complex models~(e.g., \cite{nguyen2019conflict,rossi2018altruistic,delic2018use,Tommasel2024-dp}). Additionally, neural network-based approaches require large datasets containing group decisions and feedback that are not widely available. On the other hand, social choice-based aggregation strategies offer a wide range of calculations transforming a simple matrix of individual preferences (user-item ratings) into a singular group outcome \cite{masthoff2015group,senot2010analysis}. They are often categorized as either \textit{borderline} (using only a subset of ratings, e.g., Least Misery or Most Pleasure), \textit{majority-based} (only using the most popular items/ratings, e.g., Approval Voting) or \textit{consensus-based} (using all ratings, e.g., Additive Utilitarian). In this study, we include strategies from all three categories. Included aggregation strategies are summarized in Table \ref{tab:strats}. We implement social choice-based aggregation instead of more recent approaches due to their transparent aggregation procedures and widely spread use in the literature, maximizing reproducibility.

% the usual about LLMs and (G)RS
Recently, LLMs have been implemented within the GRS pipeline for a variety of reasons. LLMs have been used as decision-making for both individual \cite{bao-etal-2024-decoding, Liao2024llara} and group recommendations \cite{Tommasel2024-dp,Waterschoot2025Consistent}. Additionally, LLMs have been used for generating explanations for (group) recommender systems \cite{Lubos2024llmgenerated,Said2025explaining,Waterschoot2025Consistent} and to address specific issues such as cold-start problems \cite{Sanner2023-kb,Wu2024could,Dai2023uncovering} or cross-domain recommendation~\cite{Petruzzelli2024instruct,Kim2024large}. A recent focus are conversational recommender systems employing an LLM as backbone model \cite{Gao2023chatrec,Yang2024behav}. While LLMs are capable of challenging the more conventional methodologies for generating recommendations, these models come with specific drawbacks such as growing computational power and privacy concerns \cite{Wu2023enhanced,Zhao2024rec,Waterschoot2025Pitfalls}. In our study, we make use of LLMs-as-judge with the goal of generating human-like assessments of perceived fairness, satisfaction and group consensus.

\begin{figure*}[h]
  \centering
  \includegraphics[width=1\linewidth]{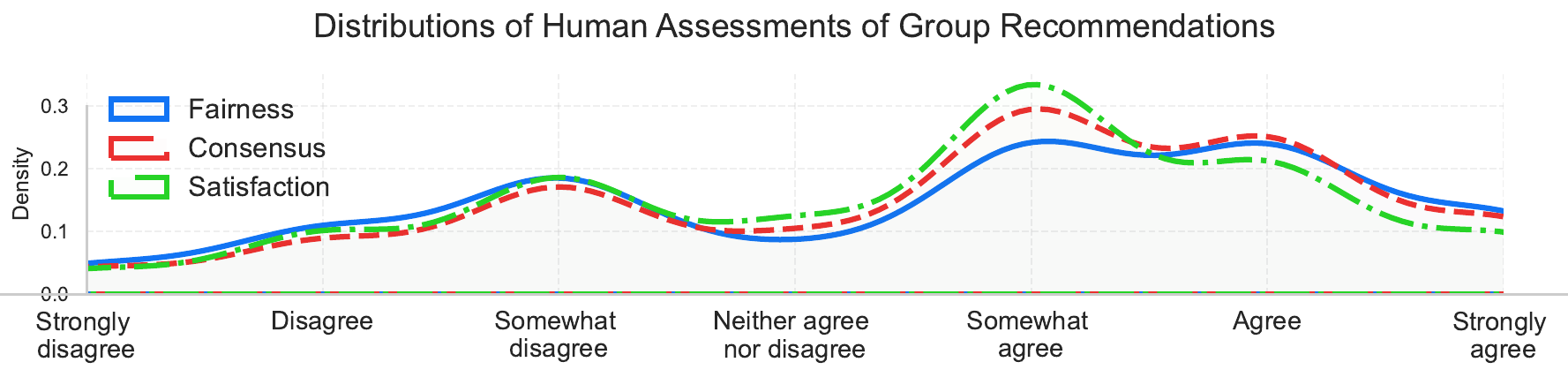}
  \caption{Human distributions of responses to rating group recommendations in terms of fairness, satisfaction and consensus on a 7-point Likert scale (n=$1,152$), collected in a the user study by \citet{barile2023evaluating}.}
  \label{fig:human}
  \Description{Human distributions from the survey by \citet{barile2023evaluating}. Data is available through \citet{barile2023data}}
\end{figure*}

\subsection{Human Assessment of GRS}\label{sec:rel-fair}
% how is fairness usually measured
Fairness is an important topic in recommender systems \cite{Deldjoo2024-cf,Kaya2020-fs,Li2023fairness}. Additional subjective assessments such as satisfaction, and specifically for group recommendation, perceived group consensus receive attention \cite{barile2023evaluating,BARILE2026Critical}. Several metrics have been proposed with regard to fairness. Each of these focus on specific aspects of fairness, such as demographic parity and equal opportunity, i.e., all qualified items are being considered \cite{Kumar2023fairness}. For GRS specifically, fairness has been defined as balancing preferences of top-ranked items across group members by \citet{Kaya2020-fs} to define a greedy algorithm to generate group recommendations that satisfy this condition.

In the context of GRS, human evaluations are usually collected through user studies \cite{BARILE2026Critical}. The literature makes the distinction between external evaluation, in which the participant is not part of the group and assessing the recommendation as an outsider \cite{barile2023evaluating,tran2019towards}, and internal evaluation in which the user provides their ratings and is part of the group itself \cite{BARILE2026Critical}. Participants are provided with a group context (user-item matrix) and are asked to rate statements on a Likert scale. In this study, we make use of the such a dataset\footnote{Source dataset can be found here: \url{https://dataverse.nl/dataset.xhtml?persistentId=doi:10.34894/8EVX4U}} \cite{barile2023evaluating,barile2023data}, which asked participants to rate \textbf{fairness} (``\textit{The group recommendation is fair to all group members}''), \textbf{satisfaction} (``\textit{The group members will be satisfied with regard to the group recommendation}''), and \textbf{consensus} (``\textit{The group members will agree on the group recommendation}''). The human response distributions from this study are visualized in Figure \ref{fig:human}.

\subsection{LLM-as-judge}\label{sec:rel-llmjudge}
The LLM-as-judge paradigm has emerged as a potential, scalable alternative to human evaluation, suited to address the traditional bottlenecks of group recruitment and the costs associated with user studies. Under this framework, LLMs are implemented as evaluators to generate human-like natural language assessments or Likert scores \cite{Bakker2022finetuning, li-etal-2025-generation}, including for subjective forms of evaluation such as sexism detection \cite{Aoyagui2025perspective}. The shift toward automated evaluation offers potential for the real-time assessment of algorithmic decision-making \cite{Baumeister2025streambased}. However, an unexplored  possibility lies in employing these models to evaluate group recommendations at scale or within a larger GRS pipeline, where complex trade-offs between individual preferences and collective satisfaction are simultaneously considered. If LLMs result in similar assessments, they can complement traditional user studies or produce real-time scoring to provide the fairest group recommendations given a specific group scenario.

Despite their promises, the performance of base models (not fine-tuned for the particular task) as impartial judges may be limited by systematic evaluative bias. \citet{shi-etal-2025-judging} demonstrate that base models often result in significant issues regarding preference fairness and position consistency. Additionally, similar to human evaluation, base models are susceptible to various forms of cognitive and social bias. \citet{malberg-etal-2025-comprehensive} provide a systematic overview of $30$ cognitive biases across $20$ state-of-the-art LLMs, while \citet{chen-etal-2024-humans} show LLM susceptibility to biases based on gender or authority, leading to skewed assessments. Furthermore, base models often demonstrate a self-preference bias, favoring output that mirrors their own style \cite{Wataoka2025-aq}. Thus, we opt to fine-tune LLMs to anchor our evaluation models in the human ground truth, evaluated using the human distribution of assessments (Figure \ref{fig:human}). We showcase the improvement when fine-tuning the LLM-as-judge models as opposed to implementing base models for evaluation.

\section{Methodology}

\begin{figure*}[h]
  \centering
  \includegraphics[width=1\linewidth]{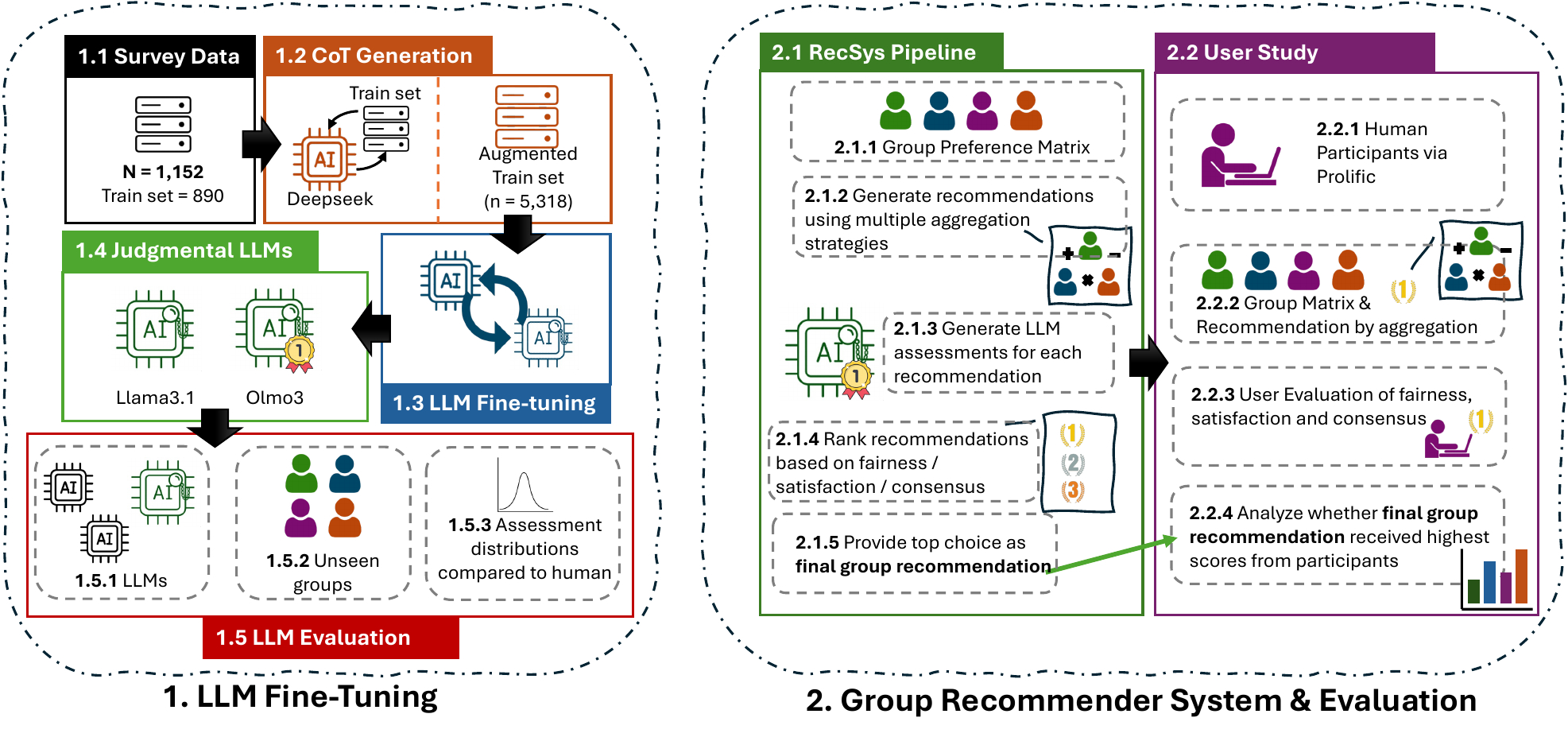}
  \caption{Visualization of the full pipeline. The methodology is divided into two main parts: (1) LLM fine-tuning for automated assessment of fairness, consensus, and satisfaction; (2) GRS, in which the best-performing, fine-tuned LLM generates assessments of potential recommendations.}
  \label{fig:pipeline}
  \Description{Full pipeline}
\end{figure*}

The following paragraphs describe the methodology used in this study. The full pipeline is visualized in Figure \ref{fig:pipeline}. First, we outline the LLM fine-tuning using an augmented training set, grounded in human ground truth. Second, we employ the best-performing model in our GRS pipeline to decide the final group recommendation based on the LLM-generated assessments. Finally, we describe the user study which we performed to evaluate the final recommendations.

\subsection{LLM Fine-Tuning}

\subsubsection{Survey Data}\label{sec:surveydata}
The starting point of obtaining fine-tuned LLMs for generating human-aligned assessments was the human ground truth presented in \citet{barile2023evaluating,barile2023data}. The authors conducted a user study in which $288$ participants rated their perception of fairness, satisfaction and group consensus based on a group preference matrix (containing ratings) and a final recommendation. A 7-point Likert scale was used and each participant rated four scenarios, resulting in a final dataset containing $1,152$ assessments. Each human assessments contained a separate score for fairness, satisfaction and consensus. The distribution of human responses is visualized in Figure \ref{fig:human}.

\subsubsection{Chain-of-Thought Generation}
To address the limited scale of the original human-annotated dataset and to provide the necessary training signals for effective fine-tuning, we employed Knowledge Distillation via Synthetic Chain-of-Thought (CoT) Generation \cite{hsieh-etal-2023-distilling}. We partitioned the data into an train/test split ($n\_training = 890$, $n\_test=262$) using stratified sampling to ensure that the score distributions remained balanced across both subsets. This approach was particularly important for capturing infrequent edge cases (the highest and lowest Likert scores) ensuring the test set remained representative of the full spectrum of human sentiment.

To expand the training set, we utilized \textit{DeepSeek-V3 (671B)} (via the \textit{Ollama} cloud API\footnote{\url{https://ollama.com/library/deepseek-v3.1}}) as a teacher model to generate diverse, synthetic reasoning for the existing human ground truth. By applying high sampling parameters ($Temperature = 0.95$,$ top\_p =0.95$), we oversampled our $890$ training instances to produce a total of $5,318$ training samples. This approach allowed us to generate multiple distinct reasoning paths for the same outcome, capturing the nuanced justifications behind identical Likert scores. For each instance, the teacher model was provided with the full recommendation scenario, the group preference table, and the specific human-assigned score. The model was then prompted to generate concise, 2-3 sentence justifications that explicitly referenced individual group members (e.g., ``\textit{Member 1 does not like this outcome...}'').\footnote{Full prompt is documented in the companion repository: \url{https://github.com/Cwaterschoot/Consensus-vs-dissent-2026/tree/main}} Throughout this expansion, we conducted checks to ensure the original human ground truth scores remained unaltered. The result of this knowledge distillation was an augmented dataset suitable for fine-tuning LLMs to generate both reasoning and assessment scores for group recommendations. The dataset is available in the companion repository.

\begin{comment}
   \textbf{only training data, test set remains the same!}
Deepseek, high sampling to generate large variety in reasoning even if human scores are identical. \textbf{IMPORTANT: human scores untouched. We oversampled human data and generated diverse reasonings for the same outcome}. Akin to model distillation, in which Deepseek was teacher model, our finetuned LLMs student model
deepseek-v3.1:671b-cloud accessible via Ollama api. Temperature 0.95, top\_p 0.95 as well. resulting in diverse reasoning for identical scores (i.e., there can be multiple reasons the person gave this outcome a 3 for fairness!) Deepseek was presented with the full scenario including the full preference table. Then we presented the scores the human gave and explained the likert scale. The LLM was instructed to generate a concise reasoning (2-3 sentences dimension) to explain the thought behind the score. Explicit references to group members were encouraged (e.g., member 1 does not like this outcome). \textbf{The human ground truth was double checked to ensure the scores were unchanged.} 
\end{comment}

\subsubsection{Fine-tuning}
We fine-tuned two open-source LLMs, \textit{Llama-3.1-8B-Instruct} and \textit{OLMo-3-7B-Instruct-SFT}, using the augmented training corpus ($n=5,318$). Both base models are available on Huggingface (Llama is open-weight, Olmo open-source). Training was conducted on a single Google Colab A100 GPU utilizing Parameter-Efficient Fine-Tuning (PEFT) via Low-Rank Adaptation (LoRA). To optimize memory efficiency, models were loaded in 4-bit NF4 quantization using bitsandbytes library. The LoRA configuration utilized a rank ($r$) of 16 and an alpha ($\alpha$) of 32. %targeting all linear layers including the attention ($q, k, v, o$) and MLP ($gate, up, down$) modules.
\footnote{Full training documentation and augmented dataset are available in the companion repository.} %\url{https://github.com/Cwaterschoot/Consensus-vs-dissent-2026/tree/main}}

A critical component of our training pipeline was the implementation of a custom label masking function. To make sure that the models learned to replicate the reasoning process and scores rather than the prompt structure and group ratings, we applied a mask to the loss function for all tokens preceding the assistant's response. This forced the gradient updates to focus exclusively on the synthetic Chain-of-Thought and the human ground truth. We employed the standard transformers Trainer (combined with our custom masking) and a learning rate of $1 \times 10^{-4}$, a batch size of 2 with 4 gradient accumulation steps.%, and Bfloat16 precision. 
The models were trained for 4 epochs, and the best-performing checkpoint was selected based on the lowest evaluation loss on the original (not augmented) test set.

%\subsubsection{Judgmental LLMs}

\subsubsection{LLM Evaluation}
To evaluate the LLMs, we make use of the exact procedure by \citet{barile2023evaluating} to generate novel group scenarios, allowing us to evaluate the LLMs with data that was not seen during fine-tuning. In total, we generated $1,180$ new group scenarios. In order to showcase the impact of the fine-tuning phase, we generated a distribution of assessments with the \textit{Llama3.1-8B} base model (not fine-tuned). 

We compared the resulting assessment distributions of the base model and both fine-tuned versions to the human ground truth using the Wasserstein Distance (Earth Mover's Distance). This metric, suitable for the ordinal Likert scale data, measures the minimum work required to transform the LLM-generated distribution into the observed human ground truth. A smaller distance (from now on referred to as WD) is better. In the results, we also report the average Likert score as well as the standard deviation to provide a full overview of the LLM-generated assessments.

\subsection{GRS and Evaluation}

\subsubsection{Recommender System Pipeline}\label{sec:recsys}
Our GRS pipeline is designed around the repeated evaluation of aggregation strategies to determine the most suitable outcome for a given group. Following the established scenario construction in GRS literature \cite{Waterschoot2025Pitfalls,barile2021toward,barile2023evaluating}, we utilize a user preference matrix containing explicit ratings for restaurants on a scale from $1$ to $5$.

Starting from the group's preferences, the system calculates the outcomes for six distinct social choice-based aggregation strategies: \textit{Additive Utilitarian} (ADD), \textit{Approval Voting} (APP), \textit{Fairness} (FAI), \textit{Least Misery} (LMS), \textit{Majority Voting} (MAJ), and \textit{Most Pleasure} (MPL). These strategies were selected due to their presence in the training data. The recommendations derived by each strategy are evaluated five separate times each, simulating the assessment of five different survey participants to ensure a variety in responses. The number of generated assessments is a configurable parameter. The final score for each strategy is determined by calculating the average across the five scores for each of the three metrics (fairness, satisfaction, consensus). Once the final scores are calculated, the strategies are ranked and the group is presented with the recommendation based on the top-ranked strategy. Our pipeline allows for deciding which of the three metrics is used as deciding factor. Consistent with the scenarios used in the original survey data \cite{barile2023evaluating, Waterschoot2025withfriends}, we account for prior group history by assuming the top three items in the resulting list have already been visited. Consequently, the system recommends the fourth-ranked item as the final output for the group. We follow the scenario structure from the literature to maximize consistency and reproducibility of GRS evaluation and data collection.

\begin{comment}
    
\textcolor{red}{Scenario: user preference matrix with user-restaurant ratings. Each user has rated each restaurant (following the scenario construction in the literature on GRS (see e.g. \cite{Waterschoot2025Pitfalls,barile2021toward,barile2023evaluating}) -> all strategies calculated -> Judgmental model assesses all possible outcomes (each of the six aggregation strategies) \textbf{EACH STRATEGY 5 times, as if assessed by 5 different people, calculate average for each metric (fairness, satisfaction, consensus) as final score of the strategy}-> strategies are ranked based on metric -> GRS recommends output based on the number 1 strategy -> group recommendation (top-1) or top x ranking to be picked from. In our case, we adopt the scenario used in the original survey data \cite{barile2023evaluating,Waterschoot2025withfriends} and create scenarios in which the top-3 have already been visited, thus recommending the 4th. Included strategies are those included in the survey data: \textit{Additive Utilitarian} (ADD), \textit{Approval Voting} (APP), \textit{Fairness} (FAI), \textit{Least Misery} (LMS), \textit{Majority voting} (MAJ), and \textit{Most Pleasure} (MPL)}
\end{comment}

\subsubsection{User Study Evaluation}
Following the code and procedure provided by \citet{barile2023evaluating} and thus, adhering to the format of the previously used survey data, we generated a set of group scenarios to be used in the user evaluation.\footnote{Our user evaluation was preregistered on OSF: \url{https://osf.io/98kuw/overview?view_only=41c006b50fb84f0abb76347c98e94ff9}} Each scenario consisted of a user-item matrix in which a group of five members is presented alongside their fictitious ratings of 10 restaurants (scale from 1 to 5). Replicating the original data, we generated group scenarios across four different group configurations (within-group preference similarity): (i) \textit{uniform} (all similar), (ii) \textit{divergent} (all dissimilar), (iii) \textit{minority} (all similar except one), and (iv) \textit{coalitional} (two subgroups) \cite{barile2023evaluating}. The scenario describes that the group has already visited three restaurants and thus, the group is now recommended the fourth restaurant in the ranked list. This list was generated by a social choice-based aggregation strategy \cite{senot2010analysis}. Included strategies are those included in the survey data: \textit{Additive Utilitarian} (ADD), \textit{Approval Voting} (APP), \textit{Fairness} (FAI), \textit{Least Misery} (LMS), \textit{Majority voting} (MAJ), and \textit{Most Pleasure} (MPL). For our specific evaluation, we add a final category to the strategy list, namely \textit{LLM}. The latter presents the participant with the recommendation generated by the strategy which was chosen by the LLM.

Participants were recruited via the online participant platform Prolific.\footnote{\url{https://www.prolific.com/}} Aggregation strategy was between-subject; each participant was assigned one of the seven options (one of six strategies or the LLM's pick) and was only presented scenarios in which the recommendations were generated by that strategy. Group configuration was within-subject, which means that every participant rated scenarios based on all four group configurations. Each participant saw one group for each group configuration (randomly selected out of four options), resulting in four groups assessed per participant. Thus, our survey contained a total of 16 group scenarios, from which the participants each saw four. This survey design matches the study used to generate the survey data described in Section \ref{sec:surveydata} and used for LLM fine-tuning. For each new group scenario, the participant had to rate the following three statements (taken from \cite{barile2023evaluating}) using a 7-point Likert scale from $-3$ (strongly disagree) to $3$ (strongly agree): ``\textit{The group recommendation is fair to all group members}'' (fairness), ``\textit{The group members will agree on the group recommendation}'' (consensus), and ``\textit{The group members will be satisfied with regard to the group recommendation}'' (satisfaction). 

In order to evaluate the performance of our GRS described in Section \ref{sec:recsys}, we constructed an mixed-design ANOVA model to test whether the \textit{LLM} category received significantly higher scores compared to the singular (static) aggregation strategies. This result would imply the benefit of dynamically adapting the aggregation strategy based on LLM-generated assessments, providing end-users with the fairest or most satisfying recommendations. Based on power analysis, we calculated a required sample size of $284$ participants. Additionally, we report average scores for fairness, satisfaction and consensus, as well as the standard deviation for each of the seven options (six strategies and our dynamic LLM-based selection). As an exploratory analysis, we compare the performance across the four group configurations. The full survey outline is found in the preregistration on OSF.\footnote{\url{https://osf.io/98kuw/overview?view_only=41c006b50fb84f0abb76347c98e94ff9}}

\section{Results} 
In the following paragraphs, we present the results of our study. First, we discuss the performance of our fine-tuned models relative to the base model. Second, we present the results of our user study performed to evaluate our proposed GRS pipeline. 

\begin{figure*}[h]
  \centering
  \includegraphics[width=1\linewidth]{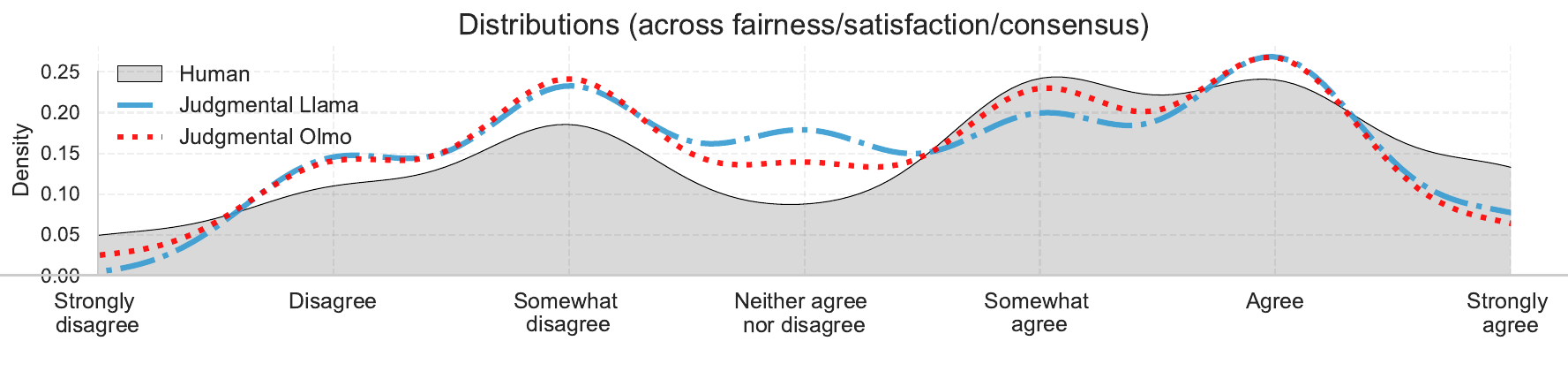}
  \caption{Distributions (across fairness, satisfaction, consensus) for both judgmental models and the human ground truth. }
  \label{fig:distr-LLMs}
  \Description{Human distributions from the survey by \citet{barile2023evaluating}. Distributions from both fine-tuned Judgmental models}
\end{figure*}

\begin{table}[htb]
  \centering
  \caption{Overview comparing the ground truth human distributions with LLM-generated assessment distributions. Models included the Llama/Olmo base models and our fine-tuned versions (Judgmental Llama and Judgmental Olmo). Lower Wasserstein Distance (WD) indicates a distribution closer to Ground Truth.}
  \label{tab:overview-metrics}
  \begin{tabular}{llccc}
    \toprule
    Assessment & Model & WD ($\downarrow$) & Mean & $\sigma$ \\
    \midrule
    \textbf{Fairness} & Llama3.1-8B (Base) & 1.57 & 1.82 & 0.54 \\
             & Olmo3-7B (Base) & 0.88 & -0.18 &1.53 \\
             & Judgmental Llama & 0.96 & 0.39 & 1.54 \\
             & Judgmental Olmo & \textbf{0.70} & 0.33 & 1.58 \\
             & \textit{Ground Truth} &  & 0.55 & 1.74 \\
    \addlinespace
    \textbf{Satisfaction} & Llama3.1-8B (Base) & 1.06 & 0.76 & 0.48 \\
                 & Olmo3-7B (Base) & 1.16 &1.29 & 1.40\\
                 & Judgmental Llama & 0.78 & 0.60 & 1.13 \\
                 & Judgmental Olmo & \textbf{0.68} & 0.31 & 1.55 \\
                 & \textit{Ground Truth} &  & 0.50 & 1.58 \\
    \addlinespace
    \textbf{Consensus} & Llama3.1-8B (Base) & 1.09 & 1.82 & 0.54 \\
              & Olmo3-7B (Base) &1.09 &0.94 &1.88 \\
              & Judgmental Llama & 0.85 & 0.10 & 1.32 \\
              & Judgmental Olmo & \textbf{0.82} & 0.15 & 1.59 \\
              & \textit{Ground Truth} &  & 0.64 & 1.63 \\
    \bottomrule
  \end{tabular}
\end{table}

\subsection{LLM Evaluation}
We evaluated the performance of our fine-tuned models (\textit{Judgmental Llama} and \textit{Judgmental Olmo}) against the base models (both Llama and Olmo) by comparing their output distributions to the human ground truth. Table \ref{tab:overview-metrics} summarizes these results across the three dimensions: fairness, satisfaction, and consensus.

The Llama base model exhibited significant misalignment with the ground truth, resulting in a high Wasserstein Distance (WD) and low standard deviation (Table \ref{tab:overview-metrics}). This suggests that the base model generated an exceedingly narrow set of assessments. In contrast, our fine-tuned models demonstrated a reduction in WD, indicating closer alignment with the human distribution (Figure \ref{fig:distr-LLMs}).%\footnote{Separate distribution plots for fairness, satisfaction and consensus are found in the repository}
While the Olmo3 base model did achieve a relatively low distance to the human distribution for fairness, its distance for the other two metrics was the highest, indicating relatively strong misalignment (Table \ref{tab:overview-metrics}).

Notably, \textbf{Judgmental Olmo} emerged as the best-performing model, achieving the lowest WD scores across all three dimensions ($0.70$ for fairness, $0.68$ for satisfaction, and $0.82$ for consensus). Additionally, the fine-tuned model replicated the variety in human responses ($\sigma > 1.55$), closely approaching the ground truth ($\sigma \approx 1.6$). Judgmental Olmo was selected as the assessment generator for the GRS pipeline and subsequent user study.

Additionally, we analyzed the Pearson correlation coefficients between fairness, satisfaction and consensus scores generated by the Judgmental Olmo model. The results show relatively strong correlations (ranging between $0.64$ and $0.78$), consistent with the strong correlations found in the original human ground truth ($0.71$ to $0.83$). From a practical perspective, these correlations imply that these factors will largely align in their ranking of recommendation outcomes, providing a consistent signal regardless of which specific dimension of the three is prioritized as a deciding factor in the GRS pipeline. For this study, we calculate the average across the three metrics and rank the aggregation strategies based on these average scores. This is the most standard approach, while future work may optimize towards either fairness, satisfaction or consensus.

\begin{comment}
\textcolor{red}{Wasserstein Distance to human distribution (show distribution plot with some base models? -> have added table already
Judgmental Olmo as the winner - so this one used as assessment generator in Section \ref{sec:res-userstudy} \textbf{results overview in Table \ref{tab:overview-metrics}}
Which strategies receive highest scores? (results skewed towards certain SCBAS?)
judgmental Olmo (best performing): 
we look at correlations because if the three factors are highly correlated, there is less argument for pickign a separate deciding factor. 
correlations of three factors between 0.64 and 0.78. So pretty correlated. So it does not really matter which one we pick to decide which recommendation to provide. In the human data, we find such strong correlations (0.71 to 0.83). So not a problem that it is correlated for judgmental Olmo
}
\end{comment}

\subsection{User Evaluation}\label{sec:res-userstudy}

\subsubsection{Participants} Our user study resulted in a total of $1,136$ assessments provided by $284$ participants. Those who failed an attention check were omitted from the analysis and were replaced by a new participant. They were required to be proficient English speakers above 18 years of age. Each participant was allowed to participate in the study once, and received a reimbursement according to Prolific guidelines.\footnote{\url{https://www.prolific.com/resources/how-much-should-you-pay-research-participants}}  Our sample consisted of $172$ ($60.5\%$) male, $105$ ($37.0\%$) female, and $7$ ($2.5\%$) non-binary participants. $101$ ($35.56\%$) were between 26-35 years old, $74$ ($26.05\%$) between 18-25, $65$ ($22.9\%$) between 36-45, $25$  ($8.8\%$) between 46-55, and $19$ ($6.7\%$) above the age of 56. 

\subsubsection{LLM Selection} To obtain the LLM category in our user study, we processed all included group scenarios using our GRS pipeline. In order to decide which strategy was dynamically selected for each given group, our recommendation pipeline generated five assessments (committee size) per potential recommendation. An assessment is a set of three scores: one for perceived fairness, satisfaction and consensus. We averaged across the five assessments per recommendation to obtain the final scores. In total, ADD was the winning strategy (picked by the LLM as best based on fairness, consensus and satisfaction scores) for six out of 16 (four groups times four configurations). FAI was the second most picked (4), while LMS and MPL were selected twice. APP and MAJ were only selected once. We saw several patterns across group configurations. For \textit{minority} groups, only ADD or FAI were selected (each strategy twice), while the MPL strategy was only used for uniform groups (alongside one occurrence of LMS and ADD). For \textit{divergent} groups, ADD was selected twice, followed by one occurrence of APP and one of FAI. Finally, for coalitional groups, each included group differed in terms of dynamically selected strategy (one times ADD, FAI, LMS, and MAJ).

\begin{comment}
chosen scenarios by system: {'LMS': 2, 'ADD': 6, 'MPL': 2, 'FAI': 4, 'APP': 1, 'MAJ': 1}

{'coalitional': {'ADD': 1, 'FAI': 1, 'LMS': 1, 'MAJ': 1},
 'divergent': {'ADD': 2, 'APP': 1, 'FAI': 1},
 'minority': {'ADD': 2, 'FAI': 2},
 'uniform': {'ADD': 1, 'LMS': 1, 'MPL': 2}}
\end{comment}
 
\begin{table}[htb]
  \centering
  \caption{Average Likert scores (-3 to 3) and standard deviation for perceived fairness, satisfaction and consensus of group recommendations. Split by aggregation strategy (between-subject).}
  \label{tab:overview-metrics-strats}
  \begin{tabular}{lccc}
    \toprule
      & Fairness & Satisfaction & Consensus\\
    \midrule
    ADD        & 0.51 (1.72) & 0.58 (1.56) & 0.76 (1.57)\\
    APP         & 0.28 (1.73) & 0.41 (1.58) & 0.47 (1.62)  \\
    FAI         & 0.61 (1.67) & \textbf{0.70} (1.64)  &0.74 (1.76)  \\
    LMS         & 0.48 (1.55) & 0.48 (1.47)  & 0.68 (1.48) \\
    MAJ        & 0.20 (1.65)& 0.20 (1.60)& 0.26 (1.71)\\
    MPL        & 0.20 (1.79)& 0.14 (1.63)& 0.18 (1.76)\\
    \midrule
    LLM         &\textbf{0.82} (1.68)& 0.68 (1.50) & \textbf{0.81} (1.56)\\
    \bottomrule
  \end{tabular}
\end{table}

\subsubsection{User Scores} Descriptive statistics reveal that our LLM-based aggregation strategy selection tends to outperformed static social choice-based aggregation (Table \ref{tab:overview-metrics-strats}), achieving the highest mean scores for perceived fairness ($0.82$) and consensus ($0.81$), and the second-highest for satisfaction ($0.68$, surpassed only by FAI, which scored $0.70$). Strategies like APP, MAJ and MPL lagged behind alternatives with regard to all three metrics. The ANOVA results for the three metrics are summarized in Table \ref{tab:anova}. P-values were adjusted using Bonferroni correction ($0.05/3=0.017$).  %perceived fairness, satisfaction and consensus of group recommendations.

\begin{table}[htb]
  \centering
  \caption{ANOVA results for perceived fairness, satisfaction, and consensus. Significance was calculated using a corrected threshold of 0.017. Statistical significance is indicated using an asterisk (*).}
  \label{tab:anova}
  \begin{tabular}{llrr}
    \toprule
    Metric & Source & $F$ & $p$ \\
    \midrule
    \textbf{Fairness} & Agg & 1.98 & .068 \\
     & Config & 49.69 & <.001* \\
     & Interaction & 5.18 & <.001* \\
    \midrule
    \textbf{Satisfaction} & Agg & 2.32 & .032 \\
     & Config & 37.81 & <.001* \\
     & Interaction & 4.28 & <.001* \\
    \midrule
    \textbf{Consensus} & Agg & 2.77 & .012* \\
     & Config & 35.96 & <.001* \\
     & Interaction & 4.49 & <.001* \\
    \bottomrule
  \end{tabular}
\end{table}

\textit{Perceived fairness.} There was no main effect for aggregation strategy with regard to \textit{perceived fairness}, however, our analysis indicated a significant interaction between group configuration and aggregation strategy ($F = 5.18, p < .001$). In a post-hoc analysis, we found a significant interaction effect in coalitional groups, with LLM outperforming both APP ($t = 4.44, p < .001$) and ADD ($t = 3.22, p = .002$) strategies.
On a global level, the LLM was perceived as significantly more fair than the Majority (MAJ) strategy ($t = 2.81, p = .006$). 

\textit{Perceived satisfaction}. The analysis revealed a significant interaction effect between group configuration and aggregation strategy ($F = 4.28, p < .001$). Our dynamic LLM aggregation selection outperformed in minority and coalitional groups. In the former, the aggregation strategies selected based on the LLM-generated assessments provided significantly higher satisfaction than both the MAJ ($t = 4.11, p < .001$) and MPL ($t = 3.01, p = .004$) strategies. Furthermore, in coalitional groups, the LLM statistically outperformed APP ($t = 3.07, p = .003$).

%consensus:
\textit{Perceived consensus}. The third and final metric was perceived consensus. Our dynamic LLM-based selection significantly outperformed MPL regardless of group configuration ($t = 2.50, p = .015$). Looking at the significant interaction effects between consensus and group configuration (Table \ref{tab:anova}), our proposed method outperformed APP ($t = 4.83, p < .001$) and ADD ($t = 3.17, p = .002$) for coalitional groups, as well as MAJ for minority configurations ($t = 3.24, p = .002$).

\begin{comment}
\textcolor{red}{less stat significant differences due to adjusted p at 0.017}
satisfaction:
group\_config * Agg     minority        MAJ  4.110228  0.000104  0.900915
group\_config * Agg  coalitional        APP  3.072021  0.002896  0.666182
group\_config * Agg     minority        MPL  3.011638  0.003652  0.659355

consensus
group\_config * Agg  coalitional        APP  4.834340  0.000006  1.047494
group\_config * Agg     minority        MAJ  3.237139  0.001784  0.710817
group\_config * Agg  coalitional        ADD  3.165676  0.002223  0.695022
Agg            -        MPL  2.497281  0.014594  0.549601
\end{comment}

\section{Discussion}
In the following paragraphs, we contextualize our results. First, we discuss that our results, both in terms of LLM evaluation and user study, strengthen the validity of the proposed fine-tuning procedure as well as its implementation for dynamic selection of aggregation strategy. Second, we outline potential adjustments and parameter tuning that we propose to investigate in future work.  Thirdly, we discuss the significant interaction with group configuration. Fourthly, we discuss potential adjustments to the pipeline. Afterwards, we discuss additional avenues for future work, including scaling down model size to optimize for real-time inference as well as comparing our pipeline to other state-of-the-art GRS methodologies. Finally, we identify and outline several limitations of our work.%\textcolor{red}{still want to add something about the interactions}

\subsection{Validity of Fine-tuning}

%Fine-tuning itself works on distilled reasoning. Base models fail. Fine-tuning is necessary. Distance to human distribution did drop as planned -> so technical part is sound, however quality data is needed (preferably with human reasoning: limitation). Pipeline already achieved highest scores for fairness and consensus, so overall concept of adapting aggregation strategy based on human-like assessments seems valid. 

Our results highlight the potential benefits of fine-tuning on human ground truth and (distilled) reasoning for aligning LLMs with human subjective assessments. We show that base models may fail to capture the nuance of group dynamics or produce biased (overtly positive) results, resulting in a narrow distribution with high Wasserstein Distance to the ground truth distribution. The fine-tuned versions successfully reduced the distance to human ground truth, allowing our implemented model to increasingly replicate the diversity ($\sigma \approx 1.6$) found in human fairness, satisfaction, and consensus ratings. The validity of our proposed pipeline is further illustrated by the user study, where the dynamic selection pipeline achieved the highest mean scores for both fairness and consensus. 

%\textbf{Clear parallels between the LLM-selected strategies and the literature, such as underperforming MPL except for uniform groups and consistent high scores for ADD and FAI \citet{barile2023evaluating}}
Additionally, many of our findings align with findings in previously conducted user studies regarding social choice-based aggregation. For example, MPL was only selected by the LLM-based pipeline for uniform groups, a result corresponding to previous findings comparing uniform groups to other configurations \cite{barile2023evaluating,BARILE2026Critical}. % While for divergent, coalitional and, to a certain extent, MPL consistently underperformed alternatives, for uniform groups, statistically significant results were found \cite{barile2023evaluating,BARILE2026Critical}. 
Additionally, these studies found consistently strong results for FAI and ADD as well, mirroring the frequency at which these strategies were selected in our pipeline. %These findings further validate our approach + dynamic adaptations in line with recommendations made based on previous user studies

By adapting the aggregation strategy based on automated, human-like assessments rather than relying on a static social choice-based aggregation strategy, the system was able to adapt to potential unfair or unsatisfying outcomes in a way that is consistent with previous human selections. These findings suggest that our dynamic approach to group recommendations,  still rooted in transparent preference aggregation, may offer a more robust and fair framework compared to traditional aggregation methods.

\subsection{The Impact of Group Configuration}
The strength of our pipeline lies not in inventing new formulas for aggregation, but in its ability to adapt when static strategies reach a recommendation that might be deemed poor by the group. 
%Our results (Section \ref{sec:res-userstudy}) showed that the LLM most frequently selected ADD and FAI as the winning strategies, a similar conclusion proposed in the literature \cite{barile2023evaluating}.
The selection of different aggregation strategies by our LLM highlights again that there are contexts where it is beneficial to deviate from what might be a globally strong strategy. For some group configurations, a strategy may produce inferior recommendations and an alternative strategy might lead to a better outcome. For example, our pipeline correctly identified that recommendations generated by MPL might be useful for obtaining the most positive assessments for uniform groups \cite{barile2023evaluating}. Additionally, our results show that the strongest strategies (ADD and FAI) might struggle with some coalitional groups, resulting in a dynamical selection in which each coalitional group received a recommendation generated from a broader pool of aggregation strategies (ADD, FAI, MAJ, LMS). Additional exploratory analysis, in which we ran $40$ additional group scenarios (10 per configuration, $n=1,200$) using our pipeline revealed that this pattern holds. MPL was strictly used for uniform scenarios (3 out of 10 uniform groups), while for coalitional groups all six strategies were chosen at least once. Similar as for the groups in oru user study, FAI was chosen most frequently.

%exploratory running: hoping to confirm patterns
% i) MPL only for uniform
% ii) ADD/FAI overall most frequent
% iii) coalitional all over the place

\subsection{Pipeline Adjustments}
Several avenues exist to refine our optimization logic within the GRS pipeline. While the current implementation made use of a simple average across fairness, satisfaction, and consensus to select the winning aggregation strategy, the system could be tuned to prioritize a single dimension depending on the specific application context. %For example, high-stake decisions might require prioritization of perceived group consensus while satisfaction might be more beneficial for more low-stake outcomes.
Furthermore, in our current study we generated five assessments (committee size) per potential recommendation outcome and calculated the average across those to obtain a final score for each metric. Future work will determine the optimal number of internal assessments, balancing robustness and computational efficiency.

While our results demonstrate that distilling reasoning from a high-capacity model like DeepSeek-V3.1 into a smaller model (such as Olmo3) is effective, this approach serves as a bridge towards an optimal outcome, i.e., training on actual human reasoning provided by participants as opposed to human-like reasoning. Additional experimentation suggests that the failure of base models is systemic \cite{Waterschoot2026fairest}. Regardless of model family or parameter size (whether 7 billion, 20 billion or 235 billion), generated assessments consistently lack behind fine-tuned output with regard to the distance to the human ground truth distribution. This finding underscores that quality fine-tuning data cannot be substituted by scale.

%\textcolor{red}{FUTURE WORK: System can still be finetuned (e.g. not taking average, but taking one metric to pick strategy (either fairness, satisfaction, consensus), committee size tuning (now we did five assessments per recommendation and took average across those 5. However this number needs to be tuned to see if a larger number of assessments produces more robust results, MORE TRAINING DATA ALSO WITH TEXTUAL REASONING). Now we distilled reasoning from a very large LLM and fine-tuned a smaller model to produce such reasoning. We can also test backbone model to fine-tune. Preliminary results indicate that in various model families are different biases, regardless of parameter size. Simply taking a very large base model without fine-tuning underperforms a smaller fine-tuned model like ours.}

\subsection{Future Work}
Our findings open up several other avenues for future work. First, while our pipeline demonstrates clear potential, the reliance on five parallel assessments per recommendation, resulting in a total of $30$ inferences per group, creates significant latency. Depending on hardware, this can go all the way up to several minutes per group. To make this judgmental framework viable within a real-time GRS pipeline, future work will explore the distillation of reasoning into Small Language Models (SLMs) with $1.5$ billion parameters or fewer (as opposed to 8B+ in the current study). Transitioning to these smaller models would drastically reduce computational requirements (e.g., GPU time) as well as overall response times, potentially allowing for a larger number of assessments per recommendation (increasing robustness) without sacrificing feasibility and user experience in real-time contexts.

Second, we strictly constrained our recommendation procedure to social choice-based aggregation strategies to prioritize transparency and take advantage of the large body of literature on explainable group recommendations. However, future research should benchmark our dynamic selection pipeline against state-of-the-art Deep Learning-based GRS models, such as attentive group recommendation frameworks \cite{cao2018attentive,chen2021attentive}. Additionally, we do not have access to the required datasets and documentation to include these approaches into our comparative analysis, as they rely on specific features or data requirements.

%\textcolor{red}{Future work: 1.5B or less model based: tiny LLMs for real-time inference of committee like assessments. Will reduce both computational requirements and latency, perfect for implementation in real-time systems. Or to increase committee size. Current setup with 5 assessments per takes too long for actual real-time inference. }

%Future work: compare with other SOTA GRS approaches (deep learning-based approaches \cite{cao2018attentive,chen2021attentive}). We explicitly limited the aggregation options to social choice-based aggregation to safeguard reproducibility and transparency (social choice-based explanations are a well-studied subject in the GRS literature (\cite{tran2019towards,barile2023evaluating,Waterschoot2025withfriends})

\subsection{Limitations}
We identified several limitations of our work. First, a primary limitation in developing our fine-tuned models is the scarcity of large-scale, high-quality human datasets that include both quantitative scores and qualitative reasoning. To circumvent this, our study utilized a distillation pipeline, generating synthetic reasoning from a high-capacity model (DeepSeek-V3.1) to fine-tune our smaller models. While our results show this to be effective, it may introduces a dependency on the teacher model. In future work, we will focus on longitudinal data collection specifically designed for model training, incentivizing participants to provide their internal reasoning in text alongside their Likert scores.

Secondly, the requirement for a five-member assessment committee per recommendation presents a computational bottleneck. In a real-time system with many end-users, the cumulative latency of generating and parsing multiple reasoning blocks hinders immediate interaction. As mentioned earlier, the use of tiny LLMs might be the key in circumventing this limitation.

Finally, our system is domain-dependent due to the scenarios present in the previously conducted user studies. Additionally, we focused solely on GRS. Whether our findings generalize beyond group recommendation specifically is unclear. For example, in our training data, the participant was presented with a group scenario in which the first three restaurants were already visited. Subsequently, the fourth-ranked restaurant became the recommendation. A more standardized input format is required to move towards domain- and scenario-independence. Due to the fact that this originates from the data collection phase, we will move towards multi-domain data collection in future work.

\begin{comment}
data bottleneck: for fine-tuning, you need large-scale, quality human data which is currently lacking. Especially if you aim to train on textual reasoning as well. We generated synthetic reasoning with large LLM (deepseekv3.1) and distilled it to our smaller models -> \textbf{future work: setting up data collection specifically tailored to gathering training data (scores + textual reasoning). ask participants to provide their own reasoning. Large enoguh data for finetuning. Usually user studies are not set up with the goal of collecting training data for thse kinds of models}

computational requirements of generating multiple sets of assessments per group (partly addressed by reducing model size etc)

Potential scenario format dependency. A universal format needs to be created in which domain can be easily adjusted. 
\end{comment}

\section{Conclusion}
%In this paper, we addressed a fundamental question for GRS: how should the recommendation pipeline be transparently adapted to the group context itself while taking into account how users perceive the outcome? More specifically, we investigated the possibility to dynamically adapt the social choice-based aggregation strategy based on the perceived fairness, satisfaction and consensus of potential recommendation outcomes. By leveraging fine-tuned LLMs, we augmented existing human evaluation data with human-like reasoning and implemented the fine-tuned models into a \textit{judgmental }framework that integrates the assessment phase, mimicking a group of human participants in a user evaluation, directly into the recommendation pipeline.

Our methodology focused on bridging the gap between LLM output and human ground truth assessments of group recommendations. By utilizing DeepSeek-V3.1 to generate reasoning rooted in ground truth, we addressed the limitations of base models. %, which exhibited a large Wasserstein distance from the human distribution. 
Through fine-tuning on this distilled reasoning and human assessments, we %reduced this distance by 
aligned the models' logic with human-like perceptions of group dynamics and recommendations. We demonstrated this by a reduction in Wasserstein distance from the human distribution. 

%This alignment allowed our best performing model to function as a proxy for human evaluators within the GRS pipeline.

We also validated the best-performing LLM approach through a user study ($n=284$). Our pipeline achieved the highest average scores for perceived fairness and consensus. Additionally, our system showed adaptability in complex groups such as the coalitional group configuration. %frequently made use of  ADD and FAI, as these strategies serve as a decent baseline for the majority of standard recommendation scenarios in terms of perceived fairness, consensus and satisfaction \cite{barile2023evaluating}. Additionally, our pipeline made use of MPL for some uniform groups, again in line with previous user study findings. 
In the case of complex scenarios, such as coalitional groups, our pipeline showed the strength of making use of the wide set of aggregation strategies, outperforming static baselines with regard to fairness, satisfaction and consensus. All in all, our pipeline shows capabilities for both groups with consensus (uniform) as dissent (coalitional).
%The system also displayed other patterns previously observed in the literature.

Moving beyond the constraints of static aggregation, our work provides a scalable blueprint for GRS that accounts for the nuanced human-like interpretation of a recommendation outcome, while retaining the transparency and range of options provided by social choice-based aggregation strategies.

%\section{Appendices}

%\subsection{Appendix}

%\begin{acks}
%This work was supported by the Faculty of Science and Engineering at Maastricht University (2024 FSE Starter Grant).

%\end{acks}

\bibliographystyle{ACM-Reference-Format}
\bibliography{reprobib}


\end{document}